%% file: neurips_2025.tex
\documentclass{article}

\pdfoutput=1

\usepackage{todonotes}
\newcommand{\name}{\texttt{VeriReason}\xspace}
\usepackage[preprint]{neurips_2025}

\usepackage{graphicx}
\usepackage{epstopdf}
\usepackage{xspace}
\usepackage{wrapfig}
\usepackage[utf8]{inputenc} 
\usepackage[T1]{fontenc}    
\usepackage{url}            
\usepackage{booktabs}       
\usepackage{amsfonts}       
\usepackage{nicefrac}       
\usepackage{microtype}      
\usepackage{xcolor}         

\usepackage{graphicx}
\usepackage{tikz}
\usetikzlibrary{automata, positioning, arrows.meta, shapes.geometric}
\usepackage{float}
\usepackage{stfloats}

\usepackage{booktabs}
\usepackage{multirow}
\usepackage{makecell}
\usepackage{colortbl}
\usepackage{threeparttable}

\usepackage{xcolor}
\definecolor{Green}{RGB}{0,128,0}
\definecolor{Red}{RGB}{128,0,0}
\definecolor{light-gray}{gray}{0.9}

\usepackage{algorithm}
\usepackage{algpseudocode}

\usepackage{microtype}
\usepackage{pifont}
\usepackage{anyfontsize}
\usepackage{xspace}

\usepackage{multicol}

\usepackage[caption=false]{subfig} 

\usepackage{pifont}
\usepackage{multirow}
\usepackage{makecell}
\usepackage[caption=false]{subfig}
\usepackage{tcolorbox}
\newtcolorbox{dialogbox}{
  arc=4mm,
  colback=blue!3,
  colframe=black,
  rounded corners,
  boxrule=0.5pt,
  fonttitle=\bfseries,
  coltitle=black,
}
\usepackage{listings}
\definecolor{codegreen}{rgb}{0,0.6,0}
\definecolor{codegray}{rgb}{0.5,0.5,0.5}
\definecolor{codepurple}{rgb}{0.58,0,0.82}
\definecolor{backcolour}{rgb}{0.95,0.95,0.92}

\usepackage{amsmath}
\usepackage{amssymb}
\usepackage{mathtools}
\usepackage{amsthm}
\usepackage{graphicx}
\usepackage{titlesec}

\usepackage{natbib}
\makeatletter
\renewcommand\NAT@force@numbers{} 
\makeatother
\setcitestyle{numbers,square,comma}
\usepackage{hyperref}       

\definecolor{color1}{rgb}{0.1,0.7,0.8} 
\definecolor{color2}{rgb}{0.9,0.1,0.1} 
\definecolor{color3}{rgb}{0.7,0.3,0.7} 
\definecolor{color4}{rgb}{0.3,0.3,0.7} 
\definecolor{color5}{RGB}{8, 102, 3} 
\definecolor{color6}{rgb}{0.53, 0.66, 0.42} 

\title{VeriReason: Reinforcement Learning with Testbench Feedback for Reasoning-Enhanced Verilog Generation}

\author{
  Yiting Wang$^{1,*}$, 
  Guoheng Sun$^{1,*}$,
  Wanghao Ye$^{1}$, 
  Gang Qu$^{1}$, 
  Ang Li$^{1,\dagger}$\\
  $^{1}$Department of Electrical Engineering, University of Maryland, Maryland, United States\\
  $^{*}$Equal contribution\\
  $^{\dagger}$Corresponding author: angliece@umd.edu
}

\begin{document}

\maketitle

\begin{abstract}

Automating Register Transfer Level (RTL) code generation using Large Language Models (LLMs) offers substantial promise for streamlining digital circuit design and reducing human effort. However, current LLM-based approaches for RTL code generation face significant challenges. Methods such as supervised fine-tuning (SFT), in-context learning, and chain-of-thought (CoT) struggle with several critical limitations in the RTL domain: the scarcity of high-quality training data, poor alignment between natural language specifications and generated code, lack of built-in verification mechanisms, and difficulty balancing between model generalization and domain specialization. Inspired by groundbreaking research such as DeepSeek-R1, which combines reinforcement learning with reasoning capabilities, we introduce \name, a comprehensive framework that integrates supervised fine-tuning with Guided Reward Proximal Optimization (GRPO) reinforcement learning specifically tailored for RTL code generation. Using our curated high-quality training examples alongside a feedback-driven reward model, \name combines testbench evaluations with structural heuristics to improve specification-code alignment and eliminate hallucinations. Iterative GRPO embeds intrinsic self-checking and reasoning capabilities, enabling the model to autonomously detect and correct functional errors. On the VerilogEval Benchmark, \name delivers significant improvements: achieving 83.1\% functional correctness on the VerilogEval Machine benchmark, substantially outperforming both comparable-sized models and much larger commercial systems like GPT-4 Turbo. Additionally, our approach demonstrates up to a 2.8× increase in first-attempt functional correctness compared to baseline methods and exhibits robust generalization to unseen designs. To our knowledge, \name represents the first system to successfully integrate explicit reasoning capabilities with reinforcement learning for Verilog generation, establishing a new state-of-the-art for automated RTL synthesis. The models and datasets are available at: 

\href{https://huggingface.co/collections/AI4EDA-CASE/verireason-682807825b20a863e758e51e}{https://huggingface.co/collections/AI4EDA-CASE}

Code is Available at: \href{https://github.com/NellyW8/VeriReason}{https://github.com/NellyW8/VeriReason}

\end{abstract}

\input{sections/intro}

\input{sections/background}
\input{sections/methodology}

\input{sections/experiments}

\input{sections/conclusion}

\bibliography{main}
\bibliographystyle{plain}


\end{document}

%% file: sections/intro.tex
\section{Introduction}

Register Transfer Level (RTL) code generation is a critical yet labor-intensive task in digital circuit design, directly impacting the efficiency, performance, and power consumption of hardware systems. Traditionally, hardware engineers manually craft RTL code using hardware description languages (HDLs) such as Verilog, which differs significantly from general-purpose programming languages due to its concurrent and structural nature. Recent advancements in large language models (LLMs) offer promising opportunities to automate RTL code generation, substantially reducing the manual effort and domain expertise required. Leveraging LLMs for RTL generation can accelerate design cycles, minimize human-induced errors, and allow engineers to focus on high-level architectural decisions rather than intricate coding details.

Despite these advantages, LLM-based RTL synthesis encounters three core challenges. First, \textbf{data scarcity}: high-quality Verilog examples—and especially paired testbenches or reasoning annotations—are rare, limiting both pretraining and supervised fine-tuning (SFT) and hampering generalization. Second, \textbf{weak natural language–code alignment}: LLMs often produce syntactically valid but functionally incorrect Verilog, misinterpreting user specifications and hallucinating invalid structural heuristics  (\textit{e.g.}, port matching, net connectivity). Third, \textbf{low first-attempt accuracy without self-checking}: current models lack intrinsic mechanisms to detect or correct their own errors, relying instead on external testbench or syntax feedback for iterative refinement. Lastly, \textbf{lack of complex logical capability:} Traditional LLMs struggle to handle the intricate interdependencies between components in hardware design, often failing to maintain consistency across module interfaces, state machines, and timing constraints. Without systematic reasoning about component relationships, models produce circuits with logical inconsistencies or incomplete implementations that meet superficial requirements but fail under comprehensive verification.

Recent advances in reasoning and reinforcement learning (RL) have introduced promising approaches to overcome these challenges. Reasoning-augmented models, such as those leveraging chain-of-thought prompting or iterative refinement, have demonstrated the ability to follow multi-step logical patterns, making them particularly suitable for hardware description languages like Verilog that require strict structural correctness and functional dependencies. These reasoning mechanisms help LLMs better understand circuit intent and adhere to design constraints, and can better ensure the alignment between natural language and result. Methods such as Guided Reward Proximal Optimization (GRPO)\cite{shao2024deepseekmath} combine the strengths of SFT with reward-driven RL, enabling models to learn effectively even with minimal data and explicit feedback. By employing RL-based strategies, LLMs are trained not merely on predicting the next token but on achieving specific, meaningful outcomes, thus improving their logical reasoning, alignment, and self-checking capabilities. 

\vspace{-10pt}
\paragraph{Our Proposed Framework.}

To address the challenges in RTL generation with LLMs, we propose a novel framework, \name, combining supervised fine-tuning (SFT) and GRPO reinforcement learning, specifically tailored for Verilog RTL generation with a specially designed dataset featuring reasoning steps and testbenches. Our approach systematically tackles four critical limitations that hinder existing LLM-based hardware design methods: data scarcity in domain-specific code, natural language-code alignment issues, lack of self-checking behavior, and insufficient complex logical capabilities. Each of these challenges requires specialized techniques that we incorporate into the \name framework, as detailed in the following sections.

\textit{Data Scarcity in Domain-Specific Code:} 
We introduce a reasoning-distillation and testbench-generation pipeline to augment existing prompt–code pairs with high-quality testbenches and human-style reasoning steps, producing a high-quality dataset. Furthermore, we demonstrate that even with as few as 20 annotated examples from the \name dataset, GRPO yields substantial performance gains, dramatically lowering the bar for required training data.

\textit{Natural Language-Code Alignment:} \name employ a reward model that evaluates generated Verilog code against specifications using feedback from structural heuristics. Through GRPO optimization, the model learns to internalize structural constraints, effectively reducing hallucinations and ensuring structural correctness by penalizing invalid constructs across both interface definitions and internal hierarchy of the circuit design.

\textit{Lack of Self-Checking Behavior:} Our reinforcement learning framework inherently encourages the model to develop self-checking capabilities by iteratively refining outputs based on testbench feedback-driven rewards. Over training iterations, the model learns to anticipate and rectify errors internally, significantly enhancing first-attempt functional correctness. 

\textit{Lack of complex logical capabilities:} \name incorporates explicit reasoning steps throughout the design process, requiring the model to articulate its design decisions and verify logical consistency before implementation. By decomposing complex circuit specifications into manageable conceptual components and reasoning about their interactions, the model develops more coherent and complete implementations that maintain logical integrity across the entire design.

Our key contributions are summarized as follows:
    \begin{itemize}

    \item We design a novel framework \name, which integrates supervised fine-tuning with GRPO-based reinforcement learning and reasoning-augmented design processes for Verilog RTL code generation.
    \end{itemize}

    \begin{itemize}

    \item Our approach addresses critical shortcomings in RTL generation through a reasoning-distillation pipeline for data scarcity, reward-driven structural evaluation for NL-code alignment, testbench feedback mechanisms for self-checking behavior, and explicit reasoning steps for handling complex logical dependencies.
    \item We create a high quality dataset with reasoning and testbench that would be open-sourced to the benefit community.
    \item The framework achieves state-of-the-art performance in RTL generation tasks, demonstrating substantial improvements in first-attempt functional correctness, structural validity, and generalization capabilities with minimal training data. It delivers up to a 2.8× increase in first-attempt functional correctness compared to baseline models while outperforming existing state-of-the-art methods across multiple benchmarks. This improvement is particularly notable in smaller parameter models. Remarkably, the framework achieves 83.1\% pass@5 on VerilogEval-Machine, even surpassing much larger models including GPT-4 Turbo.

    \end{itemize}

%% file: sections/background.tex
\section{Background}
Recent years have seen a surge of interest in applying large language models (LLMs) to hardware design, particularly for generating Register-Transfer Level (RTL) code in Verilog.\cite{liu:2023:chipnemo,liu:2024:rtlcoder,chang:2023:chipgpt,blocklove:2023:chipchat, wang2025symrtloenhancingrtlcode,thakur:2024:verigen,zhong:2023:llm4eda, yao:2024:rtllm, liu2024rtlcoderfullyopensourceefficient, tsai:2024:rtlfixer, pearce:2020:dave, fu2023gpt4aigchip}. Previous reseach have shown significant potential for LLM-based RTL generation in automating parts of the hardware design process using finetuning, or using differnt prompting techniques. However, it has also revealed several key challenges of produce correct and efficient hardware designs reliably.

\paragraph{Challenges in LLM-based RTL generation tasks}
Researchers investigated fine-tuning LLMs using domain-specific data and techniques to improve their performance~\cite{liu:2024:rtlcoder,pei:2024:betterv}. For example, Liu et.al introduced ChipNeMo \cite{liu:2023:chipnemo}, which fine-tunes a general-purpose LLM on internal NVIDIA datasets for various chip design tasks. Similarly, Thankur et.al developed VeriGen \cite{thakur:2024:verigen}  to improve Verilog generation capabilities. Subsequent works, such as RTLCoder~\cite{liu:2024:rtlcoder} is trained based on automatically generated datasets. BetterV~\cite{pei:2024:betterv}, finetunes LLM by converting Verilog code to the C language. While effective, these methods face challenges of scalability and generalizability due to their high demand for high quality instruction-code pairs. The inherent limitation of lack of real RTL code and the low quality of generated code make it hard to make further improvement. 

Moreover, LLM generated Verilog code often face the issue of hallucination. Prompt-based methods \cite{blocklove:2023:chipchat, chang:2023:chipgpt} rely heavily on the quality and clarity of the input prompts, facing difficulties in consistently aligning complex, multi-step circuit specifications with the generated code. These methods often suffer from hallucinations or syntactically correct yet functionally incorrect outputs due to inadequate contextual understanding. Moreover, they inherently lack iterative refinement capabilities, making them incapable of progressively improving RTL code quality. Chain of Thought (CoT) methods \cite{yang2025havenhallucinationmitigatedllmverilog}, which encourage models to generate step-by-step reasoning sequences, typically excel in structured reasoning tasks but face challenges in RTL contexts due to the strict requirement for functional correctness and structural precision. Although CoT enhances reasoning, its effectiveness heavily depends on the clarity and correctness of intermediate reasoning steps, which can still suffer from errors in the absence of explicit correctness feedback mechanisms. Furthermore, the CoT has made the process of generation very ineffective due to long inference time. 

\subsection{Reinforcement Learning for LLM Reasoning}

Recent research has demonstrated the potential of reinforcement learning (RL) techniques to significantly enhance the reasoning capabilities of large language models (LLMs) \cite{ouyang2022training, shao2024deepseekmath, guo2025deepseekr1}. By providing explicit rewards for logical correctness and step-wise reasoning, RL enables models to autonomously discover effective problem-solving strategies, often mirroring structured human reasoning \cite{wei2022chain, xu2025logicrl}. Applications span mathematical problem solving (where RL fine-tuning on step-by-step correctness or final answer accuracy yields substantial improvements \cite{shao2024deepseekmath, guo2025deepseekr1}) and code generation, where preference optimization and RL from feedback have led to greater code validity and efficiency \cite{chen2021evaluating}.

Most successful approaches build upon policy gradient algorithms such as Proximal Policy Optimization (PPO) \cite{schulman2017proximal} or, more recently, Group Relative Policy Optimization (GRPO) \cite{shao2024deepseekmath, lambert2024reinforcement}. GRPO, in particular, compares groups of generated responses rather than evaluating them in isolation, enabling the model to build a deeper understanding of what constitutes high-quality reasoning through relative comparisons. The effectiveness of these frameworks depends on carefully designed reward functions that accurately reflect the target domain. For hardware description tasks like Verilog generation, structural similarity as provides a clear, unambiguous reward signal \cite{wang2025largelanguagemodelverilog}, encouraging models to internalize domain-specific constraints and develop robust reasoning capabilities that translate natural language specifications into golden-code similarity.

%% file: sections/methodology.tex
\begin{figure*}[t]
    \centering
    \includegraphics[bb=0 0 554 187, width=\textwidth]{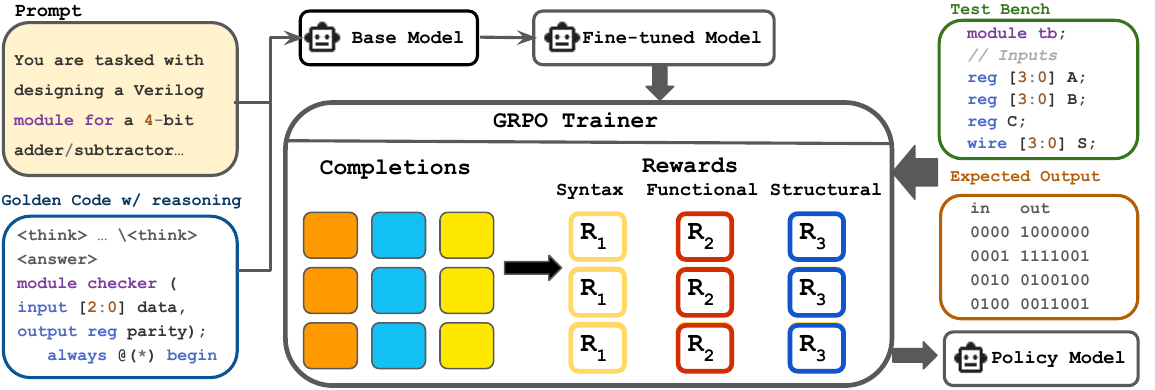}
    \caption{Workflow of \name. The framework combines supervised fine-tuning with GRPO reinforcement learning. A base model is fine-tuned and then improved through the GRPO trainer, which leverages multiple reward signals (syntax, functional, and structural correctness) derived from testbench execution and code analysis. The model incorporates explicit reasoning (<think> blocks) to break down complex hardware design tasks.}
    \label{fig:workflow}
\end{figure*}

\section{Methodology}
We present \name, a comprehensive framework that combines supervised fine-tuning with reinforcement learning specifically tailored for Verilog RTL generation. Our approach addresses four critical challenges in automated hardware design: (1) data scarcity of high-quality RTL examples, (2) weak alignment between natural language specifications and generated code, (3) lack of self-checking mechanisms in current models, and (4) insufficient complex logical reasoning capabilities for hardware design.

As shown in Figure \ref{fig:workflow}, \name employs a multi-stage approach to generate high-quality Verilog code. First, a base model is fine-tuned on a curated dataset consisting of high-quality prompt-code pairs enhanced with explicit reasoning steps and testbenches. This fine-tuned model produces initial code implementations that are then evaluated through our reward system, which combines three key components: syntax correctness, functional validation via testbench execution, and structural analysis. The GRPO (Guided Reward Proximal Optimization) trainer leverages these rewards to iteratively improve the model's ability to generate correct code while maintaining alignment with the original specification. Through this process, the model learns to incorporate reasoning steps (shown as <think> blocks) that decompose complex hardware design problems into manageable components, resulting in a policy model capable of generating functionally correct and structurally sound Verilog implementations on the first attempt.

\subsection{Reinforcement Learning Framework}
Our approach adapts Guided Reward Proximal Optimization (GRPO) specifically for RTL code generation. Unlike traditional RL methods, our framework incorporates domain-specific constraints and verification mechanisms directly into the learning process, providing immediate feedback on functional correctness, syntax, and specification adherence. This targeted optimization enables the model to efficiently learn correct Verilog implementation patterns while minimizing hallucinations and specification misalignments.

\subsubsection{Group Relative Policy Optimization (GRPO)}
We adopt Group Relative Policy Optimization (GRPO) as our core reinforcement learning algorithm due to its efficiency and demonstrated effectiveness in tasks requiring complex reasoning. GRPO provides several advantages over traditional reinforcement learning methods like Proximal Policy Optimization (PPO), including lower memory requirements and more stable training dynamics.

In GRPO, the language model serves as the policy network, taking a natural language specification $q$ as input and producing a sequence of tokens representing Verilog code as actions. The policy distribution factors across tokens: $\pi_\theta(a|q) = \prod_{t=1}^{N} \pi_\theta(a_t|q, a_{<t})$, where $\pi_\theta$ represents the policy parameterized by $\theta$, $a$ is the complete sequence of tokens (the Verilog code), and $a_t$ is the token at position $t$.

Unlike PPO, which requires a separate value function, GRPO estimates advantages using group-based sampling. For each natural language specification $q$, we generate a group of $G$ candidate Verilog implementations $\{o_1, o_2, \ldots, o_G\}$ from the current policy and compute rewards for each. The GRPO objective function is defined as:

\begin{equation}
    \mathcal{L}_{\text{GRPO}}(\theta) = \mathbb{E}_{q \sim \mathcal{D}, \{o_i\}_{i=1}^G \sim \pi_{\theta_{\text{old}}}(\cdot|q)} \left[ \frac{1}{G} \sum_{i=1}^G \min\left(r_i \cdot \rho_i, \text{clip}(\rho_i, 1-\epsilon, 1+\epsilon) \cdot r_i\right)\right] - \beta \cdot D_{\text{KL}}(\pi_\theta(\cdot|q) \| \pi_{\text{ref}}(\cdot|q)) 
\end{equation}

where:
where $\rho_i = \frac{\pi_\theta(o_i|q)}{\pi_{\theta_{\text{old}}}(o_i|q)}$ is the importance sampling ratio, $r_i$ is the normalized reward for candidate $o_i$, $\epsilon$ is a hyperparameter controlling the clipping range, $\beta$ is a coefficient balancing the KL divergence penalty, $\pi_{\text{ref}}$ is a reference policy (typically the supervised fine-tuned model), and $D_{\text{KL}}$ is the Kullback-Leibler divergence.

The advantage estimation in GRPO is simplified by normalizing rewards within each group, where $r_i = \frac{R(o_i) - \mu_R}{\sigma_R + \delta}$, with $R(o_i)$ being the raw reward for output $o_i$, $\mu_R$ and $\sigma_R$ are the mean and standard deviation of rewards within the group, and $\delta$ is a small constant for numerical stability.

This group-based normalization provides several benefits: it eliminates the need for a separate value network, reduces variance in advantage estimation, and naturally compares alternative implementations of the same specification, which aligns well with the goal of generating functionally correct Verilog code.

\subsection{Reward Model}
Our reward function combines both structural correctness and functional validation to provide comprehensive feedback during training. The reward $R$ for a generated Verilog implementation is computed as:

\begin{equation}
    R(o) = 
    \begin{cases}
        2.0, & \text{if functionally correct} \\
        0.1 + 1.0 \cdot \text{AST}_{\text{score}}(o), & \text{if syntactically correct} \\
        0, & \text{otherwise}
    \end{cases}
    \label{reward}
\end{equation}

where:
Functional correctness is determined by running the generated code through testbenches and comparing outputs with the expected behavior. Syntactic correctness is verified by successful parsing of the Verilog code. The $\text{AST}_{\text{score}}(o)$ measures structural similarity between the generated code's Abstract Syntax Tree (AST) and reference implementations, with values ranging from 0 to 1.

The AST score provides a fine-grained measure of structural correctness even when the code is not functionally perfect.
\name employs a hierarchical AST comparison algorithm specifically tailored for Verilog code structures, where $\text{ASTscore}(o) = \sum_{c \in C} w_c \cdot (0.6 \cdot \text{sim}_c + 0.5 \cdot \text{cov}_c - 0.3 \cdot \text{red}_c)$ calculates weighted structural similarity across categories $C = \{\text{module, port, always, ...}\}$ with respective importance weights $w_c$. For each category $c$, we compute sequence similarity $\text{sim}_c$ using Levenshtein distance, coverage $\text{cov}_c = |G_c \cap D_c|/|G_c|$ between generated elements $D_c$ and golden elements $G_c$, and redundancy $\text{red}_c = |D_c - G_c|/|D_c|$ to penalize hallucinated structures. This domain-specific structural analysis enables our model to maintain correct interface definitions, signal declarations, and control logic while internalizing hardware design patterns during GRPO optimization.

For functional verification, we use testbenches to evaluate the generated Verilog against reference implementations. A generated design is considered functionally correct only when it passes all test cases in the testbench, providing identical output signals to those of the golden reference for all test vectors.

\subsection{Data Preprocessing}
We address the critical issue of data scarcity in RTL generation, and the challenge of low dataset quality  through a data augmentation pipelines, and a data filtration pipeline.

\subsubsection{Data Filtration}
We implement a two-stage adaptive filtration process to optimize the dataset for GRPO training effectiveness. For each sample $s$ in our initial dataset $D$, we generate a set of $k=8$ candidate implementations $\{o_1, o_2, \ldots, o_k\}$ using our base model and compute their corresponding rewards $\{r_1, r_2, \ldots, r_k\}$ based on our reward function $R$ defined in Equation \ref{reward}.
The filtration process is formalized as follows: $D_{\text{filtered}} = \{s \in D \mid \mu_r(s) \in [\alpha_{\text{min}}, \alpha_{\text{max}}] \text{ and } \sigma_r(s) > \beta\}$, where:
$\mu_r(s) = \frac{1}{k}\sum_{i=1}^{k}r_i$ is the mean reward for sample $s$,
$\sigma_r(s) = \sqrt{\frac{1}{k}\sum_{i=1}^{k}(r_i - \mu_r(s))^2}$ is the standard deviation of rewards,
$\alpha_{\text{min}} = 0.3$ is the minimum acceptable mean reward,
$\alpha_{\text{max}} = 1.8$ is the maximum acceptable mean reward,
$\beta = 0.1$ is the minimum acceptable reward variance.
This filtration strategy excludes samples that are either too difficult (consistently low rewards) or too trivial (consistently high rewards), retaining only those samples that provide meaningful learning signals for the GRPO algorithm. Specifically, we filter out data where all rewards are zero or the average reward is below $\alpha_{\text{min}}$, as well as samples where all generations achieve near-perfect scores.

Additionally, we compute a difficulty score $\delta(s)$ for each remaining sample:

\begin{equation}
\delta(s) = 1 - \frac{\mu_r(s) - \alpha_{\text{min}}}{\alpha_{\text{max}} - \alpha_{\text{min}}}
\end{equation}

Samples are then categorized into ``simple'' and ``hard'' portions based on this difficulty score, with samples where $\delta(s) > 0.5$ classified as ``hard'' and the remainder as ``simple.'' This stratification enables targeted training strategies that progressively build model competence. The final dataset comprises 1149 samples in the hard level and 743 samples in the easy level.

\subsubsection{Reasoning Generation with Optimization}
To enhance the model's reasoning capabilities, we augment the training data with explicit reasoning steps that decompose complex hardware design problems into manageable components. Our reasoning generation pipeline extracts natural language specifications and corresponding Verilog implementations from the original dataset and uses chain-of-thought prompting with domain-specific guidance to generate detailed reasoning steps that explain the design choices, module interfaces, and implementation details. We apply an optimization process where the model is encouraged to critique its own reasoning and suggest improvements. When the model identifies a better alternative solution, we generate an improved implementation and incorporate it into the dataset. This self-improvement mechanism enables the model to iteratively refine both its reasoning process and the quality of generated code. The resulting dataset contains not only input-output pairs but also explicit reasoning traces that guide the model to develop stronger internal reasoning capabilities.

\subsubsection{Testbench Generation}
To provide reliable functional correctness signals during training, we develop an automated testbench generation pipeline. The pipeline analyzes the input-output specifications to identify signal characteristics, boundary conditions, and expected behaviors. It generates comprehensive testbenches covering both typical and edge cases, applying multiple test vector generation strategies, including directed testing for explicit requirements and constrained random testing for broader coverage. For each specification, our system generates at least 100 diverse test vectors to ensure adequate functional coverage and validates testbenches by confirming they correctly identify known-good and known-bad implementations. The generated testbenches are used both for reward computation during reinforcement learning and for final validation of model outputs. This approach ensures that the model learns to produce not just syntactically correct but functionally valid Verilog implementations aligned with the original specifications.

The combination of these data augmentation techniques with GRPO-based reinforcement learning creates a powerful framework for RTL generation that addresses the challenges of data scarcity, weak natural language-code alignment, and lack of self-checking behaviors. By encoding hardware design best practices through both structural and functional rewards, VeriReason enables the model to internalize domain-specific constraints and develop robust reasoning capabilities for RTL synthesis.

%% file: sections/experiments.tex
\section{Experiments}

\subsection{Experimental Setup}
Our primary dataset is derived from the RTLCoder \cite{liu:2024:rtlcoder} dataset of 26500 samples. We apply a thorough filtration technique on the dataset. First, we apply a simple syntax check to ensure we keep only the syntactically valid code. Then we use ChatGPT-4.1 to check whether the code matches the input prompt correctly, and generate reasoning steps for the code. For code that does not fully match the input prompt, it is re-generated and checked for syntax. We then generate a testbench for each entry with at least 100 test cases for best coverage. 

The testbench output is also saved to the dataset; in this way, during the GRPO, the testbench will only run once on the generated code, and the outcome will be directly compared to the golden code's output. Next, we run the dataset on the Qwen2.5B model to generate the code and its corresponding rewards. Based on the reward, we split the dataset into simple and hard portions for the next training steps. We end up with 1149 samples in the hard level and 743 samples in the easy level.

Our evaluation focuses on the primary benchmarks for Verilog code generation, VerilogEval \cite{liu:2023:verilogeval}. The comprehensive benchmark containing both machine-generated and human-crafted Verilog specifications. VerilogEval-Machine contains 143 samples with algorithmically generated specifications, while VerilogEval-Human includes 156 samples with human-written specifications.

We evaluate VeriReason across multiple model scales to assess parameter efficiency, including Qwen2.5-1.5B, Qwen2.5-3B, Qwen2.5-7B, and CodeLlama3-7B architectures. GRPO is used as our default RL algorithm. RL-specific settings include a generation temperature of 0.5, a total batch size of 16 (8 rollouts each), an update batch size of 2 per GRPO step with a gradient accumulation of 8, a lowered learning rate of 1.0e-6 with constant scheduler type, and repetition penalty of 1.3. The reward model follows the design in Equation \ref{reward}, with execution correctness verified using industry-standard Verilog simulators, Iverilog \cite{williams2023icarus}.

\begin{table}[t]
    \centering
    \caption{Comparative analysis of Verilog code generation performance. \colorbox{light-gray}{Gray highlighting} denotes the overall state-of-the-art results. (\textbf{bold}) indicates the best results in the model size category. Color-coded numbers show performance deltas relative to base models (\textcolor{Green}{green}: improvement).\label{tab:overall}}

    \vspace{0.25\baselineskip}
    \resizebox{1\textwidth}{!}{
        \begin{tabular}{lllccccc}
            \toprule
            \multirow[c]{2}{*}{\textbf{Category}} & \multirow[c]{2}{*}{\textbf{Method}} & \multirow[c]{2}{*}{\textbf{Params.}} & \multirow[c]{2}{*}{\textbf{Open Source}} & \multicolumn{2}{c}{\textbf{VerilogEval-Machine}} & \multicolumn{2}{c}{\textbf{VerilogEval-Human}} \\
            \cmidrule{5-6}\cmidrule{7-8}
             &  &  &  & \textbf{pass@1} & \textbf{pass@5} & \textbf{pass@1} & \textbf{pass@5} \\
            \midrule
            \multirow[c]{9}{*}{\textbf{Base Model}} & GPT-3.5-Turbo & N/A & \ding{55} & 63.5 & 78.0 & 31.2 & 47.0 \\
             & GPT-4o-mini & N/A & \ding{55} & 66.0 & 72.4 & 54.2 & 62.0 \\
             & GPT-4-Turbo & N/A & \ding{55} & 72.5 & \underline{83.0} & 64.3 & \cellcolor{light-gray}\textbf{76.1} \\
            \cmidrule{2-8}
            & Qwen-2.5-Coder & 1.5B & \checkmark & 25.6 & 40.8 & 8.3 & 17.9 \\
            & Qwen2.5-Coder & 3B & \checkmark & 48.4 & 58.9 & 21.3 & 32.7 \\
            & Qwen2.5-Coder & 7B & \checkmark & 52.7 & 69.7 & 23.9 & 41.1 \\
            & CodeLlama & 7B & \checkmark & 26.1 & 49.1 & 18.8 & 28.6 \\
             & CodeQwen1.5-7B-Chat & 7B & \checkmark & 29.1 & 61.9 & 14.8 & 36.8 \\
             & DeepSeek-Coder & 6.7B & \checkmark & 8.8 & 34.3 & 4.9 & 19.3 \\
             & DeepSeek-V3 & 671B & \checkmark & \cellcolor{light-gray}\textbf{79.2} & 80.7 & \cellcolor{light-gray}\textbf{66.1} & 72.1 \\
             
            \midrule
            \multirow[c]{6}{*}{\textbf{Fine-tuned Generation}} & ChipNeMo$^\dagger$ & 70B & \ding{55} & 53.8 & N/A & 27.6 & N/A \\
            & BetterV-CodeQwen$^\dagger$ & 7B & \ding{55} & 68.1 & 79.4 & 46.1 & 53.7 \\
            & RTLLLM$^\dagger$ & 13B & \ding{55} & 65.3 & 77.2 & 43.7 & 51.8 \\
             & VerilogEval$^\dagger$ & 16B & \ding{55} & 46.2 & 67.3 & 28.8 & 45.9 \\
            \cmidrule{2-8}
             & VeriGen$^\dagger$ & 16B & \checkmark & 44.0 & 52.6 & 30.3 & 43.9 \\
             & RTLCoder-DeepSeek-Coder & 6.7B & \checkmark & $\text{37.2}_{\textcolor{Green}{\text{+28.4}}}$ & $\text{64.9}_{\textcolor{Green}{\text{+30.6}}}$ & $\text{16.9}_{\textcolor{Green}{\text{+12.0}}}$ & $\text{35.7}_{\textcolor{Green}{\text{+16.4}}}$ \\
            \midrule
            \multirow[c]{4}{*}{\textbf{VeriReason (Ours)}} & VeriReason-Qwen2.5-1.5B & 1.5B & \checkmark & $\textbf{44.7}_{\textcolor{Green}{\text{+19.1}}}$ & $\textbf{49.1}_{\textcolor{Green}{\text{+8.3}}}$ & $\textbf{23.5}_{\textcolor{Green}{\text{+15.2}}}$ & $\textbf{26.7}_{\textcolor{Green}{\text{+8.8}}}$ \\
             & VeriReason-Qwen2.5-3B & 3B & \checkmark & $\textbf{55.9}_{\textcolor{Green}{\text{+7.5}}}$ & $\textbf{72.8}_{\textcolor{Green}{\text{+13.9}}}$ & $\textbf{33.2}_{\textcolor{Green}{\text{+11.9}}}$ & $\textbf{47.4}_{\textcolor{Green}{\text{+14.7}}}$ \\
             & VeriReason-Qwen2.5-7B & 7B & \checkmark & $\textbf{69.8}_{\textcolor{Green}{\text{+17.1}}}$ & \cellcolor{light-gray}\textbf{$\text{83.1}_{\textcolor{Green}{\text{+13.4}}}$} & $\textbf{47.9}_{\textcolor{Green}{\text{+24.0}}}$ & $\textbf{58.4}_{\textcolor{Green}{\text{+17.3}}}$ \\
             & VeriReason-codeLlama-7B & 7B & \checkmark & $\text{51.3}_{\textcolor{Green}{\text{+25.2}}}$ & $\text{64.0}_{\textcolor{Green}{\text{+14.9}}}$ & $\text{27.5}_{\textcolor{Green}{\text{+8.7}}}$ & $\text{39.9}_{\textcolor{Green}{\text{+11.3}}}$ \\
       \bottomrule
            \multicolumn{8}{l}{$^\dagger$: Reported Results.}\\
        \end{tabular}
    }
\end{table}

\subsection{Main Results}

\noindent Table~\ref{tab:overall} compares \name against state-of-the-art Verilog generation models. Our approach achieves superior performance across all model sizes, with VeriReason-Qwen2.5-7B demonstrating remarkable gains over base models: +17.1 and +24.0 percentage points on pass@1 for VerilogEval-Machine and VerilogEval-Human, respectively.

\noindent Even our smallest model, VeriReason-Qwen2.5-1.5B, outperforms many larger models after applying our reinforcement learning framework. The performance gains are particularly notable for smaller models, demonstrating the effectiveness of our approach in optimizing parameter efficiency. We observe that VeriReason-Qwen2.5-7B achieves state-of-the-art performance on Machine pass@5 (83.1\%), outperforming even GPT-4-Turbo on this metric, despite having significantly less parameters.

\noindent Furthermore, all VeriReason models establish themselves as the \textbf{best performers} in their respective parameter size categories (1.5B, 3B, and 7B), highlighting the robustness of our approach across model scales. The substantial improvements observed in VeriReason-codeLlama-7B (+25.2 percentage points in Machine pass@1) further demonstrates that our method generalizes effectively across different model architectures.

\subsection{Training Dynamics Analysis}

\begin{figure*}[h]
    \centering
    \includegraphics[bb=0 0 534 213, width=0.9\textwidth]{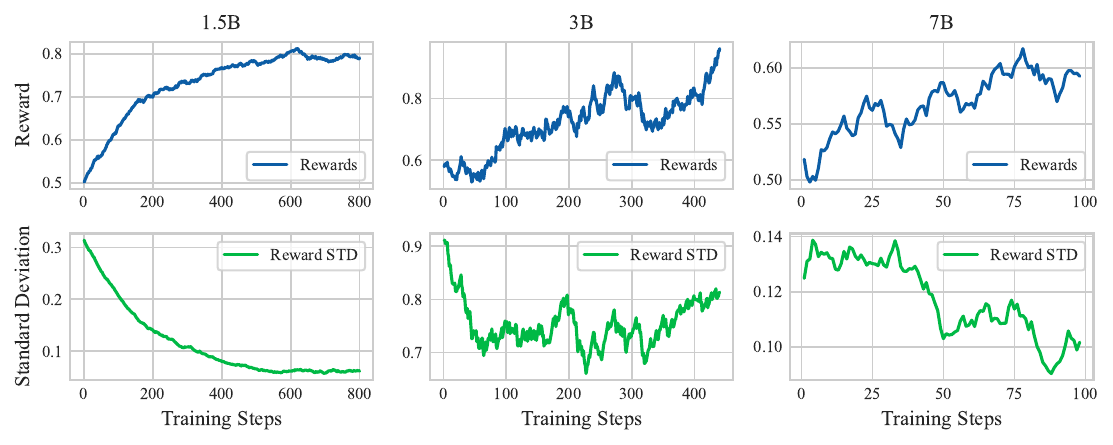}

    \caption{Reward line and std line of \name}
    \label{fig:rewards}
\end{figure*}

Figure~\ref{fig:rewards} illustrates the training dynamics of our VeriReason models across three different parameter scales (1.5B, 3B, and 7B). We track both the mean reward values (top row) and their standard deviations (bottom row) throughout the reinforcement learning process.

\subsubsection{Reward Progression}
The reward curves reveal distinct learning patterns across model sizes. The 1.5B model demonstrates steady, monotonic improvement in reward values from approximately 0.5 to 0.8 over 800 training steps, suggesting a consistent optimization path. In contrast, the 3B model exhibits higher variance in its learning trajectory, with reward values fluctuating between 0.6 and 0.8, before ultimately converging to above 0.8 by step 400. The 7B model shows the most pronounced oscillatory behavior, with rewards ranging between 0.50 and 0.65, reflecting the increased complexity of optimizing larger parameter spaces.

\subsubsection{Reward Stability}
The standard deviation plots (bottom row) provide critical insights into training stability. The 1.5B model demonstrates exceptional convergence properties, with reward variability consistently decreasing from 0.3 to below 0.1 throughout training. This smooth reduction in standard deviation correlates with the steady increase in mean rewards, indicating robust learning. The 3B model presents a more complex pattern, with initial rapid variability reduction followed by fluctuations between 0.7 and 0.8, suggesting periodic exploration-exploitation transitions. The 7B model's standard deviation exhibits the most dynamic behavior, oscillating between 0.09 and 0.14, which aligns with its more variable reward progression.

Interestingly, despite having fewer parameters, the 1.5B model achieves the most stable convergence pattern, with monotonically decreasing standard deviation. This suggests that smaller models may benefit more consistently from our reinforcement learning framework, while larger models engage in more extensive exploration of the parameter space before convergence. The final standard deviation values (approximately 0.05 for 1.5B, 0.8 for 3B, and 0.1 for 7B) indicate that all models eventually reach stable policy configurations, though with different convergence trajectories.

This analysis provides empirical evidence that our reinforcement learning approach effectively optimizes models across different parameter scales, with larger models requiring more complex optimization paths but ultimately achieving higher reward values, consistent with their superior performance on the VerilogEval benchmarks shown in Table~\ref{tab:overall}.
\vspace{-5pt}
\subsubsection{The effect of SFT and GRPO}
\begin{wraptable}{r}{0.5\textwidth} 
\centering
\caption{Ablation studies results.}
\label{tab:ablation}
\vspace{0.25\baselineskip}
\resizebox{0.5\textwidth}{!}{%
\begin{tabular}{llcccc}
\toprule
\multirow{2}{*}{\textbf{Model}} & \multirow{2}{*}{\textbf{Training Stage}} & \multicolumn{2}{c}{\textbf{VerilogEval-Machine}} & \multicolumn{2}{c}{\textbf{VerilogEval-Human}} \\
\cmidrule(lr){3-4}\cmidrule(lr){5-6}
& & \textbf{pass@1} & \textbf{pass@5} & \textbf{pass@1} & \textbf{pass@5} \\
\midrule
\multirow{3}{*}{\textbf{Qwen2.5-1.5B}} & Base & 25.6 & 40.8 & 8.2 & 17.9 \\
& + SFT & 38.6 & 46.3 & 17.8 & 23.9 \\
& + GRPO & \textbf{44.7} & \textbf{49.1} & \textbf{23.5} & \textbf{26.7} \\
\midrule
\multirow{3}{*}{\textbf{Qwen2.5-3B}} & Base & 48.4 & 58.9 & 21.3 & 32.7 \\
& + SFT & 51.9 & 69.9 & 31.3 & 45.1 \\
& + GRPO & \textbf{55.9} & \textbf{72.8} & \textbf{33.2} & \textbf{47.4} \\
\midrule
\multirow{3}{*}{\textbf{Qwen2.5-7B}} & Base & 52.7 & 69.7 & 23.9 & 41.1 \\
& + SFT & 63.4 & 79.9 & 43.4 & 56.2 \\
& + GRPO & \textbf{69.8} & \textbf{83.1} & \textbf{47.9} & \textbf{58.4} \\
\midrule
\multirow{3}{*}{\textbf{CodeLlama-7B}} & Base & 26.1 & 49.1 & 18.8 & 28.6 \\
& + SFT & 41.1 & 58.3 & 23.2 & 31.5 \\
& + GRPO  & \textbf{51.3} & \textbf{64.0} & \textbf{27.5} & \textbf{39.9} \\
\bottomrule
\end{tabular}
}
\end{wraptable}
Table \ref{tab:ablation} presents a systematic analysis of how each training component—Supervised Fine-Tuning (SFT) and Group Relative Policy Optimization (GRPO)—contributes to VeriReason's performance across different model sizes. The results demonstrate that SFT provides substantial initial performance gains across all model architectures, with smaller models showing the most dramatic relative improvements. This suggests that SFT effectively addresses the challenge of limited domain knowledge by providing high-quality training examples with explicit reasoning steps.
The addition of GRPO further enhances performance across all models and benchmarks, demonstrating the value of reinforcement learning with testbench feedback. This synergistic relationship shows that while SFT provides the foundation for understanding Verilog syntax and semantics, GRPO enhances the model's ability to produce functionally correct implementations by internalizing structural constraints and developing self-checking capabilities.

%% file: sections/conclusion.tex
\vspace{-5pt}

\section{Conclusion}
\vspace{-5pt}
\noindent This paper presents \name, a comprehensive framework integrating supervised fine-tuning with GRPO-based reinforcement learning for Verilog RTL generation. By combining explicit reasoning with testbench-driven feedback, our approach addresses key challenges in LLM-based hardware design: data scarcity, weak language-code alignment, lack of self-checking behavior, and insufficient logical reasoning. While \name achieves state-of-the-art performance across model scales, the approach incurs significant computational overhead during both training (requiring numerous testbench evaluations per iteration) and inference (where reasoning steps increase generation time by 2.5-3×). Despite these limitations, the consistent improvements across different architectures demonstrate our method's robustness and transferability. Our reward model's integration of structural correctness and functional validation encourages models to develop intrinsic self-checking capabilities—a crucial advancement for autonomous hardware design. By open-sourcing our models and datasets, we aim to accelerate progress in LLM-based hardware design, transform digital circuit development practices, and inspire future work on computational efficiency.